\documentclass[10pt,letterpaper]{article}
\usepackage{times}
\usepackage{nameref,hyperref}
\usepackage[aboveskip=1pt,labelfont=bf,labelsep=period,singlelinecheck=off]{caption}
\usepackage{lastpage,fancyhdr,graphicx}
\usepackage{epstopdf}
\fancyheadoffset[L]{2.25in}
\fancyfootoffset[L]{2.25in}

\usepackage{url}
\usepackage[numbers]{natbib}
\usepackage{amsmath,MnSymbol,mathtools}

\usepackage[]{graphicx}
\chardef\bslash=`\\ 

\newcommand{\argmax}{\mathop{\mathrm{arg\,max}}}
\usepackage{color}
 \usepackage{float}

\begin{document}

\vspace*{0.35in}

\begin{flushleft}
{\Large
\textbf\newline{Distributed Multivariate Regression Modeling For Selecting Biomarkers Under Data Protection Constraints}
}
\newline
\\
Daniela Z\"oller\textsuperscript{1,2,*},
Harald Binder\textsuperscript{1,2}
\\
\bigskip
\bf{1} Institute of Medical Biometry and Statistics, Faculty of Medicine and Medical Center - University of Freiburg,
Stefan-Meier-Str. 26, 79104 Freiburg, Germany
\\
\bigskip
\bf{2} Freiburg Center for Data Analysis and Modelling, University of Freiburg, Germany
\\
\bigskip
* daniela.zoeller@uniklinik-freiburg.de

\end{flushleft}

\section*{SUMMARY}
The discovery of clinical biomarkers requires large patient cohorts and is aided by a pooled data approach across institutions. In many countries, data protection constraints, especially in the clinical environment, forbid the exchange of individual-level data between different research institutes, impeding the conduct of a joint analyses. To circumvent this problem, only non-disclosive aggregated data is exchanged, which is often done manually and requires explicit permission before transfer, i.e., the number of data calls and the amount of data should be limited. This does not allow for more complex tasks such as variable selection, as only simple aggregated summary statistics are typically transferred. Other methods have been proposed that require more complex aggregated data or use input data perturbation, but these methods can either not deal with a high number of biomarkers or lose information. Here, we propose a multivariable regression approach for identifying biomarkers by automatic variable selection based on aggregated data in iterative calls, which can be implemented under data protection constraints. The approach can be used to jointly analyze data distributed across several locations. To minimize the amount of transferred data and the number of calls, we also provide a heuristic variant of the approach. When performing global data standardization, the proposed method yields the same results as pooled individual-level data analysis. In a simulation study, the information loss introduced by local standardization is seen to be minimal. In a typical scenario, the heuristic decreases the number of data calls from more than 10 to 3, rendering manual data releases feasible. To make our approach widely available for application, we provide an implementation of the heuristic version incorporated in the DataSHIELD framework.\\
$ $\\
Keywords: Boosting; Distributed Learning; Data Protection Constraints; Multivariable Prognostic Score; DataSHIELD
\hspace*{-4pc} {\small\it }\\[1pc]
\noindent\hspace*{-4.2pc}






\section*{Highlights}
\begin{itemize}
\item Introduction of a new distributed version of likelihood based boosting for regression modeling
\item Automatic variable selection of biomarkers in a multivariate approach
\item Introduction of a heuristic to ensure data parsimony
\item Implementation within the DataSHIELD framework
\end{itemize}

\section*{In brief}
In real-world applications, the number of potential biomarkers is often very high and an automated variable selection procedure for regression models is needed. One avaialbe appraoch is regularized regression based on componentwise likelihood-based boosting. We propose distributed version of this approach. We also introduce a heuristic version which reduces the amount of shared information to ensure data parsimony.

\section*{Introduction}
\subsection*{Motivation}
There are many potential advantages of pooling data from several sources, such as in meta-analysis of clinical studies and epidemiological cohorts (e.g. \cite{Byrne2018,Schulz2019}), the joint analysis of routine data from several hospitals (e.g. \cite{Prokosch2018,Hinderer2017}), or the combined analysis of genomic data from several cohorts and several data types \cite{Shafi2019}. Beside an increase in statistical power, more complex statistical approaches can be used when the sample size is large, e.g. for selecting the most important markers from a large set of candidate markers for outcome prediction. A joint model across several source populations might also be more reliable.

Ideally, one would pool the data on the individual-level, but this is often not possible or desirable. Several reasons might hinder sharing, such as governance restrictions, e.g., concerning data protection, fear of loss of intellectual property \cite{Fecher2015}, or logistic hurdles \cite{Gaye2014a}. Although there is a recent discussion on how to alleviate such problems as consent and confidentiality when sharing data, e.g., \cite{Wegscheider2017, Williams2017}, e.g., by introducing a broad consent as in Germany \cite{Zenker2022},  pooling of individual-level data will be problematic for years to come.  Frequently, standard meta-analysis techniques are used, which rely on aggregated summary statistics that no longer reveal individual-level information, and therefore can be exchanged and pooled for joint results across data sets. One example is the MIRACUM (Medical Informatics for Research and Care in University Medicine) consortium \cite{Prokosch2018}, in which ten German university hospitals are working on joint analysis of their data, but the data must remain within each site as in general no consent has been given to share the individual-level data with other sites. However, these standard meta-analysis approaches do not allow for more complex tasks, such as selecting important biomarkers in regression modeling with a large number of candidate biomarkers, as is needed in settings with genomic measurements or extensive characterization of cancer patients in routine data. 

\subsection*{Proposed method}

We propose a regularized regression approach based on componentwise likelihood-based boosting \cite{Tutz2006a, Tutz2007} that can perform automatic variable selection in an iterative fashion while only requiring aggregated statistics. Thus, we are able to re-formulate the underlying mathematical calculations of a classical machine learning approach in such a way that we do not need direct access to individual-level data. More precisely, the approach is based on univariable effect estimates obtained from linear regression for the outcome of interest, and pairwise covariances of the potential markers. The method can be applied to continuous and binary outcomes, and can potentially be adapted to (competing) event times via pseudo values \cite{Andersen2010,Graw2009,Zoller2015a}. We further propose a heuristic version of the algorithm to reduce the required number of covariances.  Although there is no way yet of reconstructing individual-level data from covariances in a high-dimensional setting, a reduction of shared information reduces the risk for the future as the same was believed for gradients until recently proven wrong \cite{Zhu2019}.  The heuristic is furthermore combined with an approach for calling covariances in blocks to reduce the number of data calls, potentially rendering manual data release feasible. In addition, the approach can be used at a single site to grant data protected analysis access meaning that the scientist will only see non-disclosive aggregated data and the final result.

\subsection*{Related work}

Some approaches already exist for obtaining regression models from data distributed across several sites under data protection constraints, but these are frequently limited in the number of candidate markers that can be taken into account.  For example, He et al. proposed a sparse meta-analysis algorithm for high-dimensional data solely based on aggregated data \cite{He2016a}. The algorithm uses ordinary least-squares estimates, and thus the number of potential covariates needs to be smaller than or equal to the number of participants. Other methods implement data protection by input perturbation \cite{Simmons2016a,leng2022federated}, meaning that there is an information loss beforehand. Jones et al. proposed a generalized linear model based on aggregated data which yields the same result as the model based on individual-level data \cite{Jones2012}. Lu et al. developed a similar approach in the setting of survival data for the Cox model \cite{Lu2015}, and Emam et al. derived the same result for logistic regression \cite{Emam2013}. Similar approaches have also been implemented in the DataSHIELD software framework \cite{Gaye2014a}, which enables straightforward use in projects. Unfortunately, all of these approaches have in common that they cannot deal with the situation where the number of markers is larger than the total number of samples over all sites. While the method proposed by Li et al. can handle this kind of data in regularized logistic regression models \cite{Li2016}, no automated variable selection to obtain a small set of potentially important markers is provided.  Alternative methods reformulate the problem into a separable form, but offer no garuentees for privacy protection \cite{5496140}.  Often, univariable meta-regression approaches are used. It has been shown that penalized multivariable regression approaches outperform single biomarker analysis while simultaneously selecting a sparse set of biomarkers (including most of the informative ones) and handling correlated data \cite{Ayers2010}. In general, sparse multivariable regression approaches have gained importance in the field of genomic data analysis \cite{Zhu2017,Liu2007}. To the best of our knowledge, however, none of these approaches have yet been transferred to distributed data settings. 

\subsection*{Structure}
This paper is organized as follows. In section Methods, we introduce an approach for automated selection of markers in regression models for distributed data and its heuristic variant, and a simulation design for subsequent investigation. An evaluation is presented in the Results section, with a focus on variable selection, prediction performance, and number and size of data calls.

\section*{Methods}
\label{sec_methods}
\subsection*{Stagewise regression} 
\label{sec_stage}

Let $y_i$ denote the observed outcome and $x_{i.}=\left(x_{i1},\ldots , x_{ip}\right)'$ the $p$-dimensional vector of covariates for individual $i$, $i=1,\ldots ,n$. The notation $x_{.j}$ will be used to indicate vectors $x_{.j}=\left(x_{1j},\ldots ,x_{nj}\right)'$, $j=1,\ldots , p$, containing the covariate information across all individuals for covariate $j$.

We will use the following multivariable linear regression model:
$$y_i=\beta_0+ x_{i.}' \beta + \epsilon,\ i=1,\ldots,n$$
with $\epsilon_i \sim N\left(0,\sigma^2\right)$ and $\beta=\left(\beta_1, \ldots, \beta_p\right)'$ the $p$-dimensional parameter vector. Note that this model will be misspecified when dealing with other response types, e.g., a binary response, but we will nevertheless use it. In particular, the model might produce predicted values higher than 1 or lower than 0, but it is still useful for classification, hypothesis testing and, as in our case, variable selection \cite{Pohlman2003}. To achieve the latter and also to allow for high-dimensional settings with $p \gg n$, a regularized regression approach will be used to estimate $\beta$. Regularized, or penalized, multivariable regression approaches have been shown to outperform single biomarker selection. In general, the most informative biomarkers are selected automatically while simultaneously the number of included biomarkers is kept low \cite{Ayers2010}.

We specifically choose componentwise likelihood-based boosting \cite{Tutz2006a, Tutz2007} as a regularized regression approach, which can be adapted for estimation with distributed data as will be shown below. Componentwise likelihood-based boosting is a stagewise regression approach \cite{Efron2004}, which can construct a stable risk score with respect to the outcome even in the presence of a large number of (potentially highly correlated) covariates \cite{Binder2013, Hastie2007}. The approach belongs to the class of stagewise regression approaches and yields the same results as the more widely used LASSO method if the data are orthogonal, but the variable selection is more stable for correlated data \cite{Hastie2007}.  The basic idea is to initially set the estimated parameter vector to $\hat{\beta}=(0,\ldots,0)'$, and to update this vector in a stagewise manner. In each of a potentially large number of steps $M$, one determines the covariate which most improves the model fit, using regularized estimation of candidate models and score statistics. Only the corresponding element of the estimated parameter vector $\hat\beta$ is updated, keeping all other elements fixed. 

In the following, we will to assume that the outcome is centered, i.e., $1/n \sum_i y_i = 0$, and that the covariates are standardized such that $x_{.j}'x_{.j}=n-1, j=1,\ldots,p$, which is typical for regularized regression approaches. The detailed algorithm then is as follows:

\begin{enumerate}
    \item     Initialize the estimated parameter vector $\hat{\beta}^{(1)}$ to $\hat{\beta}^{(1)}=(0,\ldots,0)'$, and the offset $\hat\eta_i^{(1)}$ to $\hat{\eta}_i^{\left(1\right)}=0$, $i=1,\ldots, n$. Set the shrinkage parameter $\nu$ to some small positive value, e.g., $\nu=0.1$.
    \item At each boosting step $m=1,\ldots,M$:
    \begin{enumerate}
        \item Consider the $j=1,\ldots,p$ potential candidate models 
        $$y_i = \hat{\eta}_i^{\left(m\right)} + x_{ij} \gamma_j^{\left(m+1\right)}  + \epsilon_i,\ i=1,\ldots,n,$$\\
        where the offset $\hat{\eta}_i^{(m)}$ incorporates the information from the previous boosting steps (more details in step 2.e).\\
        For each covariate $j$, the least-squares estimator is given by $$\hat{\gamma}_j^{\left(m+1\right)} :=  \frac{1}{n-1} \sum_{i=1}^n x_{ij} \left(y_i - \hat{y}_i^{(m)}\right)$$ with $\hat{y}_i^{(m)}$ given by $\hat{y}_i^{(m)}=x_{i.}'\hat{\beta}^{(m)}$.
        \item Calculate $\frac{\left(S_j^{\left(m+1\right)}\right)^2}{n-1}$ , $j=1,\ldots,p$, where $$S_j^{\left(m+1\right)} :=  \sum_{i=1}^n x_{ij} \left(y_i - \hat{y}_i^{(m)}\right) \propto \widehat{\gamma_j}^{\left(m+1\right)}$$ is the score function, and determine the index $j^\ast$ by $$\argmax_{j=1,\ldots,p}\left(\left(S_j^{\left(m+1\right)}\right)^2\right).$$
        \item Set $\overline{\gamma_j}^{\left(m+1\right)} :=  \nu \widehat{\gamma_j}^{\left(m+1\right)}$.
        \item Update the estimated parameter vector:
        $$\widehat{\beta_j}^{\left(m+1\right)}=
        \begin{cases}
            \hat{\beta_j}^{\left(m\right)} + \overline{\gamma_j}^{\left(m+1\right)},& \text{if } j=j^\ast\\
            \hat{\beta_j}^{\left(m\right)},                                            & \text{else.}
        \end{cases}
        $$
        \item Update the offset:
        $$\hat{\eta_i}^{\left(m+1\right)}=x_{i.}'\hat{\beta_j}^{\left(m+1\right)},\ i=1,\ldots,n$$
    \end{enumerate}
\end{enumerate}

\subsection*{Distributed Boosting}
Intuitively, each site contributes to the score statistic independently of the other sites. The score statistic is a measure of the influence of adding a single variable to the existing model and describes the model improvement. By calculating the non-disclosive contribution separately for each site and combining the contribution in a central calculation, we are able to obtain the score statistic without direct access to the individual-level data. 

\begin{figure}[H]
\centering
\includegraphics[width=\textwidth]{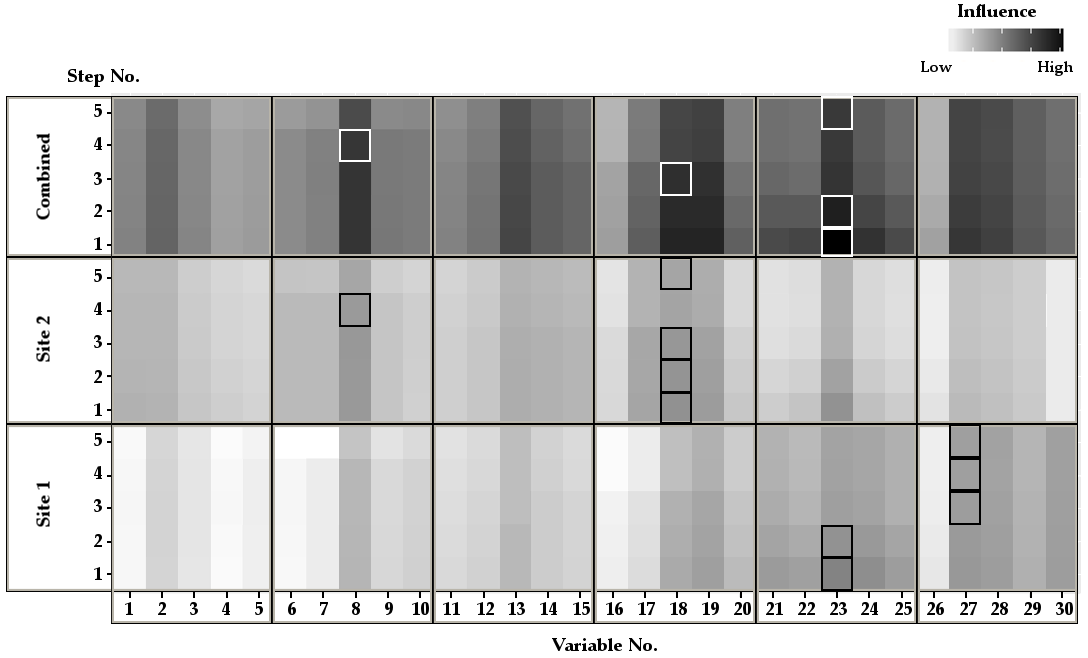}
\caption{Demonstration of the first 5 steps of the distributed boosting approach for data distributed over two sites with a binary outcome. On the x-axis, 30 out of a total of 2500 correlated variables are shown, of which only the variables 3, 8, 13, 18, 23, 28, 33, 38, 43, and 48 affect the binary outcome. The correlation is higher between the 5 variables in each box. On the y-axis, the estimated influence of selecting each variable into the model at the next step is shown separately for the two sites and for the combination, whereas the values for the sites are estimated given the previous selection of the combined method.  The darker the coloring, the higher the influence of including the variable at the next step and thus the greater the model improvement. The variables with the highest influence at each step are marked with small boxes separately for the combined method and the sites, whereas small white boxes represent the actual selected variables in the distributed boosting approach and small black boxes the variables which would have been selected at each step if only one site was available. The underlying data are simulated according to the simulation design described in Section \nameref{sec_metheva}, with 250 individuals per site and one strong effect bearing variable per higher correlated group.}
\label{Figure_Demonstration}
\end{figure}

An example of the algorithm is illustrated in Figure \ref{Figure_Demonstration}. The data are simulated; more details can be found in Section \nameref{sec_metheva}. We assume that the data are distributed across two sites with a sample size of $250$ per site. In total, we simulated $2500$ variables, of which only the variables 3, 8, 13, 18, 23, 28, 33, 38, 43, and 48 affect the binary outcome. Neighboring variables have a low correlation overall and a strong correlation in groups of five. At the beginning of the distributed boosting approach, the model contains no variables and we calculate the influence, i.e. the score statistic, of including each of the $2500$ variables per site. If only one site is available, the variables 23 or 18, respectively, are selected as these variables improve the model fit the most. When we combine the score statistics of the two sites via the distributed boosting method, the above variables still have a high estimated influence, but variable 23 has the highest. Thus, we include variable 23 in the model by estimating its effect via the maximum likelihood method as in standard regression approaches. To avoid overfitting, we use a shrunken estimate rather than the full estimate, e.g., multiplied by $0.1$. Consequently, we might underestimate the effect of variable 23, but if this is the case, the algorithm will again select the same variable at a later step. In the example, this happens at the second boosting step. At both steps, the score statistic of the complete block of highly correlated variables, i.e. variables 21 to 25, become smaller. This demonstrates that variables highly correlated with an effect bearing variable appear to influence the outcome of interest and that the boosting approach selects the one from this group with the highest estimated influence. This is also true for the two blocks directly placed beside this block as these variables are also correlated with variable 23, though to a lesser extent. At the third boosting step, variable 27, a variable with no effect on the outcome, has the highest estimated influence for site 1, whereas for site 2 this is still variable 18. By combining the two sites with our approach, variable 18 is correctly selected. Thus, either site can be most influential in the combined selection. At boosting step 4, site 1 still wrongly selects variable 27, but site 2 now switches to variable 8, another effect bearing variable. In the combination, variable 8 has the highest score and is selected. Although we would select variable 27 at the following boosting step for site 1 and variable 18 for site 2, variable 23 has the highest estimated influence in the combination. Thus, by combining the information from several sites we can gain additional information.

In the following, we will focus on the mathematical formulation to show how the algorithm introduced in Section \nameref{sec_stage} can be used with distributed data or without access to the individual-level data. It is important to note that the score function $S_j^{\left(m+1\right)}$, which is the basis for the score statistic and is used to decide which variable should be selected next, can be re-written as 
\begin{equation} 
S_j^{\left(m+1\right)} = \sum_{i=1}^n x_{ij} \left(y_i - \hat{y_i}^{\left(m\right)}\right) = \sum_{i=1}^n x_{ij} y_i - \sum_{k:\vert\hat{\beta_k}^{\left(m\right)}\vert > 0} \hat{\beta_k}^{\left(m\right)} \sum_{i=1}^n x_{ij} x_{ik}
\end{equation}
This means that only the terms $\sum_{i=1}^n x_{ij} y_i$ and $\sum_{i=1}^n x_{ij} x_{ik}$ are needed for the calculation. These are aggregated across individuals, and can be passed on without revealing individual-level information. Taking into account centering and standardization as specified above, $\sum_{i=1}^n x_{ij} y_i$ corresponds to the univariable regression coefficients, and $\sum_{i=1}^n x_{ij} x_{ik}$ corresponds to pairwise covariances up to a constant factor. Thus, individual-level information is only needed at the beginning to calculate the univariable effect estimates and the pairwise covariances, and this calculation can be separated from the algorithm itself. 

By sharing the above-stated summary statistics, it is possible to grant access to the information contained in the data without having to grant access to the individual-level data, permitting a joint analysis across several sites. Assume that the individuals are distributed over $L$ sites with $n_l$ being the sample size at site $l$, $l=1,\ldots,L$, such that $\sum_{l=1}^L n_l = n$. We propose re-writing formula (1) in the following way
\begin{equation}
	\begin{split}
		S_j^{(m+1)} &= \sum_{l=1}^L \left(\sum_{i_l=1}^{n_l} x_{i_l j} y_{i_l}
			- \sum_{k:|\widehat{\beta}_k^{(m)}|>0} \widehat{\beta}_k^{(m)}  \sum_{i_l=1}^{n_l} x_{i_lj}x_{i_lk}\right)\\
			&= \sum_{l=1}^L \sum_{i_l=1}^{n_l} x_{i_l j} y_{i_l}
			- \sum_{k:|\widehat{\beta}_k^{(m)}|>0} \widehat{\beta}_k^{(m)} \sum_{l=1}^L \sum_{i_l=1}^{n_l} x_{i_lj}x_{i_lk}
	\end{split}
\end{equation}

Consequently, the univariable regression coefficients and the pairwise covariances can be calculated for each data site and pooled afterwards without losing information. 

Above we assumed centering of the outcome and standardization of the covariates. From the mathematical point of view, this would ideally be done based on a pooled calculation of the mean and the variance, but it will probably be more convenient to perform this separately for each site, both with respect to calculation effort and data protection. This will increase the standard error of the estimates, but as we are mostly interested in variable selection we will later evaluate the effect of this on the performance of our proposed method. Alternatively, one could also calculate the global mean and subsequently the global variance to perform global centering and standardization, requiring a high number of aggregated data transfers. It is up to the scientist to decide whether the additional calculation effort is worthwhile and whether the data needed for global centering and standardization can be shared for all variables.

The number of boosting steps can be fixed by the user either by setting the number of boosting steps to the maximum number of covariates to be included in the risk score, or by selecting the boosting step with the desired number of included covariates. Another possibility would be to determine the number of boosting steps by cross-validation based on the individual data at the biggest data site. In the following, we will only consider the first possibility of a fixed number of covariates included in the model.

\subsection*{Heuristic Distributed Boosting}

The approach proposed above does not require transfer of the complete $p \times p$ matrix of pairwise covariances, as only the covariances between variables already included in the model at each boosting step and all other variables are needed. Therefore, only a few rows of the covariance matrix need to be called from the data sites at each specific boosting step. In general, the number of covariates included in the model is rather small, meaning that only a small number of rows of the whole covariance matrix will be transferred, but these rows might be very long in a high-dimensional setting. To further reduce the number of covariances to be called, one can use a heuristic that requires only a subset of the covariance matrix rows at each boosting step. In addition, the number of boosting steps at which data has to be called from the sites is decreased by adding a block data call approach. The details are given in the following.

We propose using a heuristic version of componentwise likelihood-based boosting introduced in \cite{Binder2013}. The underlying idea is that the score statistics, which are used to determine the updates at each boosting step, typically decrease or at least stay the same from one boosting step to the next. Only in very rare situations, where highly correlated variables have opposing directions of effects, is the heuristic unreliable. \cite{Binder2013} propose calculating new values of the score statistics at boosting step $m$ only for covariates with indices
$$\lbrace j:\hat{\beta}_j^{\left(m-1\right)} \neq 0\rbrace \cup \lbrace j: S_j^{\left(1\right)} \geq \min_{l:\hat{\beta}_l^{\left(m-1\right)} \neq 0} S_l^{\left(m\right)}\rbrace .
$$ 
Consequently, we need to calculate the actual score statistic for each covariate at the first boosting step, which is obtained via the univariable effect estimate. Subsequently, only a small fraction of the rows of the covariance matrix has to be transferred at each step. 

To reduce the number of data calls, which might even render manual releases of data feasible, we consider an alternative version of the heuristic approach. Instead of calling only single values of the matrix of covariances, a whole block of covariances is called which are most likely to be needed at later boosting steps. Specifically, we increase the candidate set by the $w$ covariates with the highest score statistics $S_j^{\left(1\right)}$ and $S_j^{\left(1\right)} < \min_{l:{\hat{\beta_l}}^{\left(m-1\right)} \neq 0} S_l^{\left(m\right)}$ and call the complete covariance matrix of these covariates.

\subsection*{Implementation in the DataSHIELD framework}
\label{sec_framework}

\begin{figure}[H]
\centering
\includegraphics[width=0.6\textwidth]{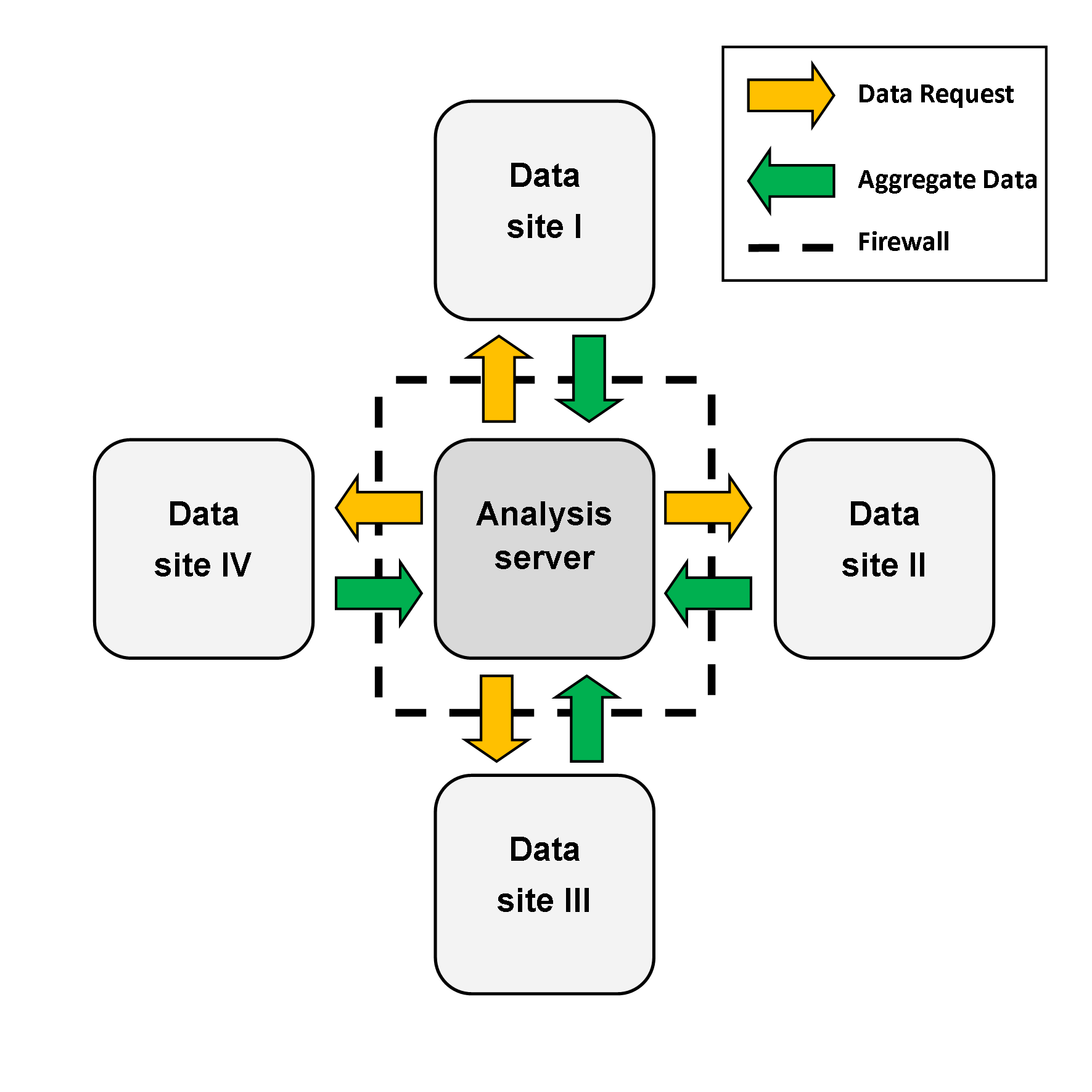}
\caption{Proposed exemplary system structure to handle communication and data flow between the participating sites. The analysis server can request aggregated data from the data sites. If the data request is in line with the data protection constraints and the restrictions agreed by the participating parties, they calculate the aggregated data and send them back. The analysis server has no access to the individual-level data and the data sites are protected via firewalls.}
\label{Figure_ITstructure}
\end{figure}

Figure \ref{Figure_ITstructure} illustrates the proposed system structure for performing the analysis, which is the same as in the DataSHIELD project (see for example \cite{Gaye2014a}). The analysis server has no direct access to the individual data and can only request aggregated data needed for the algorithm from the data sites, which could potentially be released and sent manually by each data site. The data sites do not share any information between each other and communicate only with the analysis server. 

We have incorporated the version of our new algorithm into the DataSHIELD software framework, where the data needed for the (heuristic) distributed boosting can be obtained using already implemented DataSHIELD functions - \texttt{ds.glm()} for the univariable effect estimates and \texttt{ds.cov()} for the pairwise covariances.  Correspondingly, we provide a readily accessible implementation of the proposed heuristic distributed boosting algorithm at \texttt{http://github.com/danielazoeller/ds\_DistributedBoosting.jl}. Due to the current implementation of DataSHIELD, manual release by the data sites is not yet possible.

\subsection*{Simulation Design} \label{sec_metheva}

We evaluate the method using a simulation study which mimics the data structure as found, e.g., in genome-wide association studies, as this is one of the fields where the proposed approach for distributed identification of markers might be useful. Specifically, we simulate a large number of strongly correlated variables. As such structures might also be expected in other applications, such as analysis of routine data from several clinics, the simulation study results will also be relevant beyond a genome-wide association study setting. We focus on a binary endpoint, e.g., representing response yes/no, instead of a continuous one as we have shown that the method leads to exact results for the latter even with a large number of sites, and a simulation study is thus only needed to evaluate the effect of the mis-specification.

For $n=500$ or $n=1000$ individuals, respectively, $p=2500$ covariates with values $-1$, $0$ or $1$ were generated. In a first scenario, the covariates are moderately correlated, where covariates with neighboring indices have the same value with a probability of $0.5$. In a second and third scenario, we additionally group the covariates into sets of $5$ with larger correlations. Within these groups, the probability of having the same value for two neighboring covariates is $0.75$, while the probability remains $0.5$ outside and between the groups. From the $p=2500$, only $10$ or $50$ covariates, respectively, affect the binomially distributed endpoint, e.g., representing responses yes/no. The other covariates have an effect size of $0$, and the method should thus exclude them from a model for the response. The effect size of the $10$ or $50$ covariates with an effect is $1$ or $0.2$, respectively. There are either one or two true non-zero effects per more strongly correlated group. We distribute the effects over neighboring correlation groups, leaving at least two variables without an effect between every two variables with an effect. We evaluate the effect of the number of data sites by splitting the data set into either $1$, $2$, $5$, $10$ or $20$ equally sized cohorts (or datasets). In total, we created $1000$ independent simulation datasets per simulation setting. 

We evaluate the results of the variable selection, the prediction performance with regard to the outcome, the number and the size of data calls. The main focus of the analysis is on the variable selection. In clinical practice, variables selected by a statistical method are validated, often using very elaborate experimental designs. If the variable selection is insufficient, limited resources are wasted. The variable selection is evaluated using the mean proportion of true positive selections of effect bearing covariates (i.e. non-zero effects) and false positive selected covariates overall. To determine the mean proportion of correctly selected covariates (i.e. true positives), we calculated the proportion of selection for each effect bearing covariate across all simulation runs and averaged these values across all effect bearing covariates. To obtain the mean proportion of covariates falsely selected (i.e. false positives), we calculated the mean proportion of covariates with no effect from the first $10$ selected covariates. The prediction performance is evaluated using the AUC. The aim here is to evaluate if the distribution of the data over an increasing number of sites decreases the prediction performance of the final model compared to an analysis using pooled individual-level data. In both cases, we used the first $10$ selected covariates, as a way to compare different approaches for a number of selected markers which might realistically be further explored, e.g., through validation steps, in real applications.

We compare the variable selection results of the proposed approach to the variable selection results based on a standard univariable approach, where univariable effect estimates are obtained using logistic regression models for each data site. These effect estimates are then combined using a fixed-effects meta-analytic model and the $10$ covariates with the smallest p-values are regarded as successfully selected. Additionally, we contrast the results to those obtained using the boosting approach based on a correctly specified model concerning the endpoint (`Individual data - Binomial response' \cite{Tutz2006a}) to evaluate the influence of the misspecification resulting from having a binary rather than a continuous response. The results of the boosting approach based on the individual data (`Individual data - Gaussian response'  \cite{Tutz2006a}) and based on the distributed, but globally standardized data, are also considered to verify that the proposed new approach does not lose information due to the use of aggregated and distributed data.

The main aim of the simulation study is to quantify the influence of the standardization mechanism by contrasting the results obtained after standardization of each data site to results obtained after a global standardization. In addition, we evaluate the classical and the block version of the heuristic distributed boosting method by measuring the number of called covariance values and the number of data calls needed for additional aggregated data in the above-described simulation scenarios. For this purpose, we calculate the distribution of the combination of these two quantities over $100$ boosting steps and $100$ independent simulations.

The simulation study was performed using the statistical environment R (version 3.4.4) with the \texttt{GAMBoost} package (version 1.2-3.) and the \texttt{AUC} package (version 0.3.0). For the proposed new approach, we used our implementation in the Julia language (version 0.4.7 and version 1.1.3).

\section*{Results}
\label{sec_results}
\subsection*{Evaluation of variable selection}
\label{sec_varsel}
\begin{figure}[H]
\centering
\includegraphics[width=\textwidth]{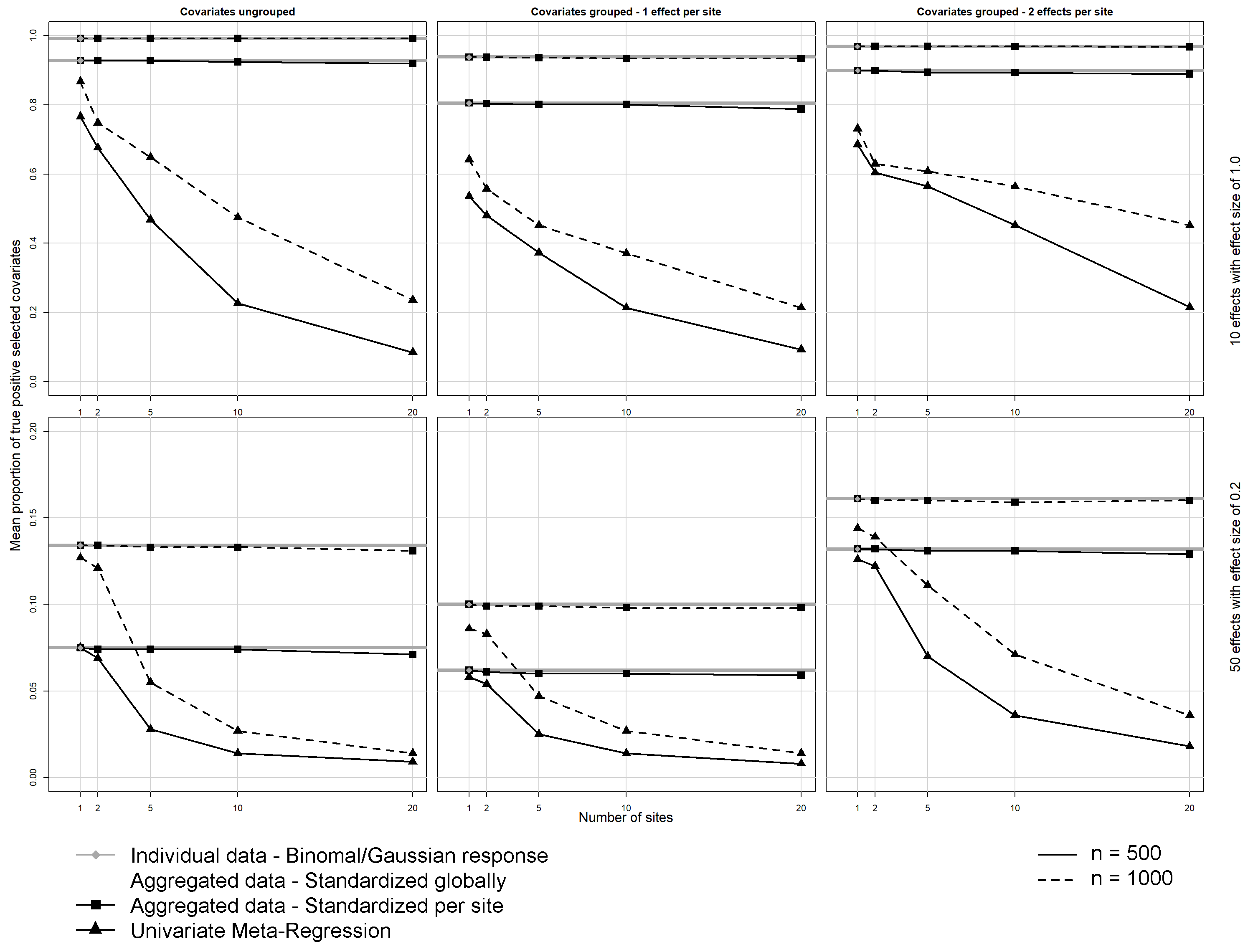}
\caption{Mean proportion of true positive selection of effect bearing covariates for distributed data and different numbers of cohorts for the proposed method (boosting based on aggregated data) compared to univariable meta-regression and boosting based on individual patient data. The results are averages over $1000$ simulated data sets per simulation setting. For each effect bearing covariate, the inclusion frequency was calculated and averaged over all simulation runs in one setting. Larger values correspond to a better true positive selection. The settings differ in the correlation structure (ungrouped covariates with a moderate correlation overall vs. grouped covariates in groups of 5 with a higher correlation), the number of effects per correlation group (1 vs 2), and the number of effect bearing covariates and effect sizes (10 covariates with an effect size of $1.0$ vs 50 covariates with an effect size of $0.2$). All other covariates have an effect size of $0.0$.  The sample size is $n=500$ (solid line) or $n=1000$ (dashed line), respectively. The results using the individual-level data (Binomial or Gaussian response) and the aggregated data with a global standardization are equal and are represented by the grey line. The results for the model with a Gaussian response using pooled individual patient data standardized locally are represented by squares. As a comparison, univariable meta-regression results are given (triangles).}
\label{Figure_TPS}
\end{figure}

In Figure \ref{Figure_TPS}, we present the mean proportion of true positives with respect to variable selection. Larger values correspond to better detection of covariates with true effects. The performance loss introduced by the misspecified endpoint is seen to be negligible in all scenarios. If one increases the number of boosting steps and thus the number of included covariates, the difference becomes somewhat larger, but overall the difference is small. In our experience, the maximum difference between the results obtained using the individual data and the correctly specified model (``Individual data – Binomial response'') and the individual data and the misspecified model (``Individual data – Gaussian response'') in the boosting approach is about $0.01$ and thus very small (results not shown). As expected, the results using individual data based boosting (Gaussian response) and aggregated and potentially distributed data based boosting with a global standardization (Gaussian response) are equal, as basically the same method is used. Thus, the proposed method is a valid alternative to already established methods when one cannot pool the individual patient data.

If the number of cohorts, consortium partners, or data sites is smaller than $10$, the results of the new method in combination with a local standardization only differ slightly from the results obtained with pooled individual data. Increasing the number of cohorts can decrease performance as the sample size becomes too small to achieve good estimates for the standardization process, but the decrease is only minimal even with only $25$ individuals per data site and a high number of effect bearing covariates with small effects.

Additionally, the new proposed method is seen to outperform univariable meta-regression considering the $10$ covariates with the smallest p-values in all scenarios. The performance of the standard approach is comparable to the newly proposed method only if the number of individuals is large and there is only one data site. The difference between the methods becomes more apparent when the number of data sites is large.

\begin{figure}[H]
\centering
\includegraphics[width=\textwidth]{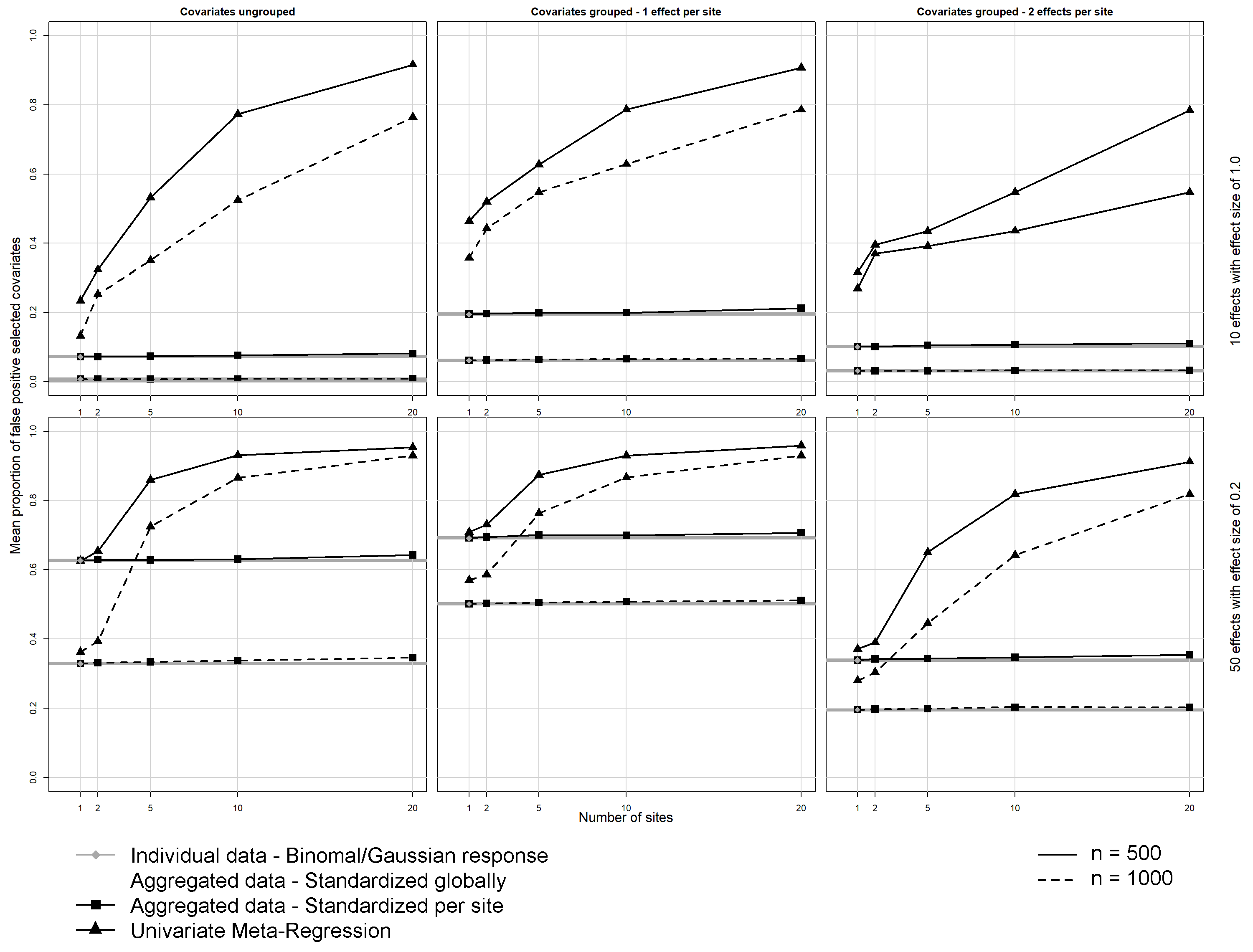}
\caption{Mean proportion of false positive selected covariates for distributed data and different numbers of cohorts for the proposed method (boosting based on aggregated data) compared to univariable meta-regression and boosting based on individual patient data.  The results are averages over $1000$ simulated data sets per simulation setting. For each simulation run, the proportion of false positive selected covariates is calculated for the first 10 selected covariates and averaged over one setting. Lower values correspond to a better false positive selection. The settings differ in the correlation structure (ungrouped covariates with a moderate correlation overall vs. grouped covariates in groups of 5 with a higher correlation), the number of effects per correlation group (1 vs. 2), and the number of effect bearing covariates and effect sizes (10 covariates with an effect size of $1.0$ vs. 50 covariates with an effect size of $0.2$). All other covariates have an effect size of $0.0$.  The sample size is $n=500$ (solid line) or $n=1000$ (dashed line), respectively. The results using the individual-level data (Binomial or Gaussian response) and the aggregated data with a global standardization are equal and are represented by the grey line. The results for the model with a Gaussian response using pooled individual patient data standardized locally are represented by squares. As a comparison, univariable meta-regression results are given (triangles).}
\label{Figure_FPS}
\end{figure}  

A similar performance pattern is seen when considering the mean proportion of false positives with respect to variable selection, as displayed in Figure \ref{Figure_FPS}. A large value corresponds to poor variable selection performance. In the simulation scenarios with $10$ effect bearing covariates, on average less than $30\%$ of the $10$ selected covariates are truly covariates with no effect. In settings with a sample size of $1000$, the proportion is even lower and does not exceed $20\%$. Distributing the data over two cohorts is again seen to have only a minimal impact, and the results are comparable to those obtained with the individual data. In particular, there is no increase in the average number of falsely selected covariates.

\subsection*{Prediction performance}
\label{sec_perf}
\begin{figure}[H]
\centering
\includegraphics[width=\textwidth]{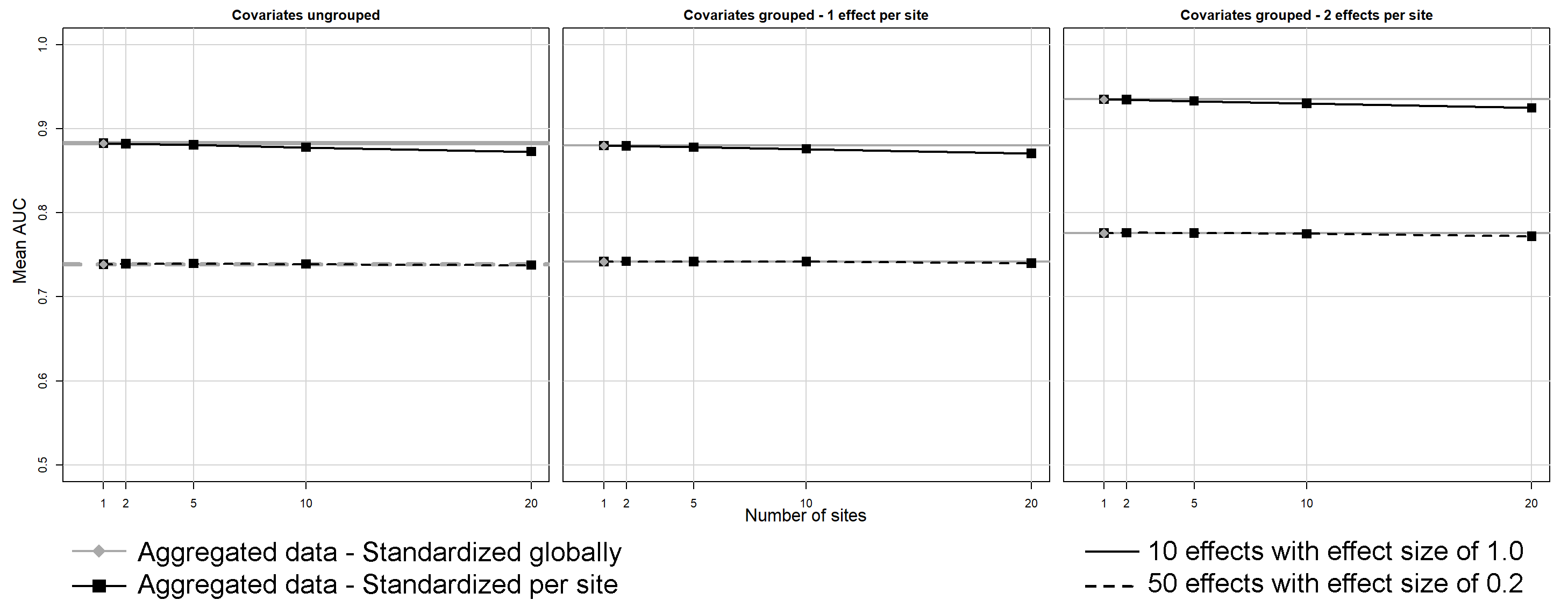}
\caption{Mean AUC for distributed boosting using locally standardized data on distributed data for different numbers of sites. The results are based on $1000$ simulated data sets per setting. For each simulation run, the AUC is calculated and averaged over one setting. Higher values above 0.5 correspond to a better classification and 0.5 to a random classification. The settings differ in the correlation structure (ungrouped covariates with a moderate correlation overall vs. grouped covariates in groups of 5 with a higher correlation), the number of effects per correlation group (1 vs. 2), and the number of effect bearing covariates and effect sizes (10 covariates with an effect size of $1.0$ (solid lines) vs. 50 covariates with an effect size of $0.2$ (dashed lines)). All other covariates have an effect size of $0.0$. The sample size is $n=500$. The results for the globally standardized distributed boosting approach are shown by a grey line and are equal to what one would obtain using boosting on individual-level data. The results for the locally standardized distributed boosting approach are shown by black squares.}
\label{Figure_AUC}
\end{figure}

In Figure \ref{Figure_AUC}, we present the results for prediction performance of the proposed distributed boosting approach. Overall, the mean AUC is above $0.85$ in the setting with $10$ strong effects and above $0.70$ in the setting with $50$ weak effects. The best prediction performance is obtained in the scenario with strongly correlated covariates and $2$ covariates per strongly correlated covariate group. Thus, even if the boosting algorithm selects a covariate with no true effect, the wrongly selected covariate can explain some of the variance if this covariate is strongly correlated with an unselected covariate with an effect. If we distribute the data over a large number of cohorts and standardize the data locally, the AUC is only minimally decreased; otherwise we do not observe a decrease in the AUC. Consequently, and in combination with the results for the variable selection, the distribution does not significantly affect the size of the effect estimates.

\subsection*{Number and size of data calls}
In the following, we compare the number of data calls and their size between the classical and the block version of the proposed heuristic distributed boosting approach, in different scenarios with varying buffer sizes. We measure the data call size by the number of called covariances and the size of the data calls by the number of times the algorithm requires new data.

The minimum number of data calls is two: one data call for the univariable effect estimates to obtain the score statistics before the first boosting step, and one data call for the first required covariances. In the standard distributed boosting approach, the algorithm requires one row of the covariance matrix per data call. Thus, taking into account symmetry and a known diagonal due to standardized covariates, we would call $\left(p-\text{Number of covariates included in the model}\right)$ covariances in each data call, independent of the covariance structure. We initialize a new data call if the algorithm adds a new covariate to the model, meaning that at worst one would need the same number of data calls as boosting steps. In the given simulation setting with $p=2500$ and a maximum of $100$ boosting steps, the number of values needed in each data call / boosting step for covariances ranges from $2400$ to $2499$. In the evaluation described above, we considered the first $10$ included covariates. In total, we required $29 945$ covariate values called in $10$ data calls to obtain these models. The heuristic and the block-heuristic versions aim to reduce both the number of required covariates as well as the number of data calls.

\begin{figure}[H]
\centering
\includegraphics[width=\textwidth]{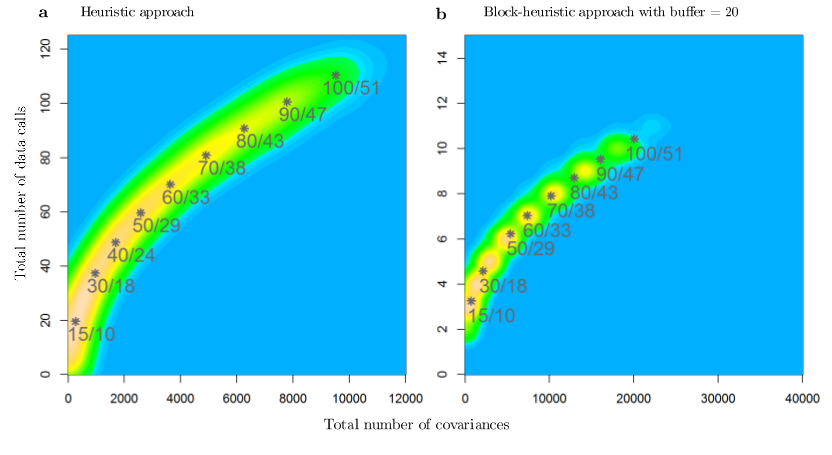}
\caption{Comparison of heuristic (panel a) vs. block-heuristic (panel b) version with buffer = 20 of in a scenario with $50$ weak effects bearing covariates and a moderate overall covariance. The distribution of the total number of called covariances vs. the total number of data calls is calculated marginally over $100$ boosting steps and $100$ simulations. A blue color corresponds to zero, and the more yellow the color, the more often the specific combination occurs. The grey stars represent the mean number of data calls and called covariances for specific boosting step numbers (before the slash) and corresponding model sizes (after the slash).}
\label{Figure_heuristic}
\end{figure}

\begin{figure}[H]
\centering
\includegraphics[height=0.82\textheight]{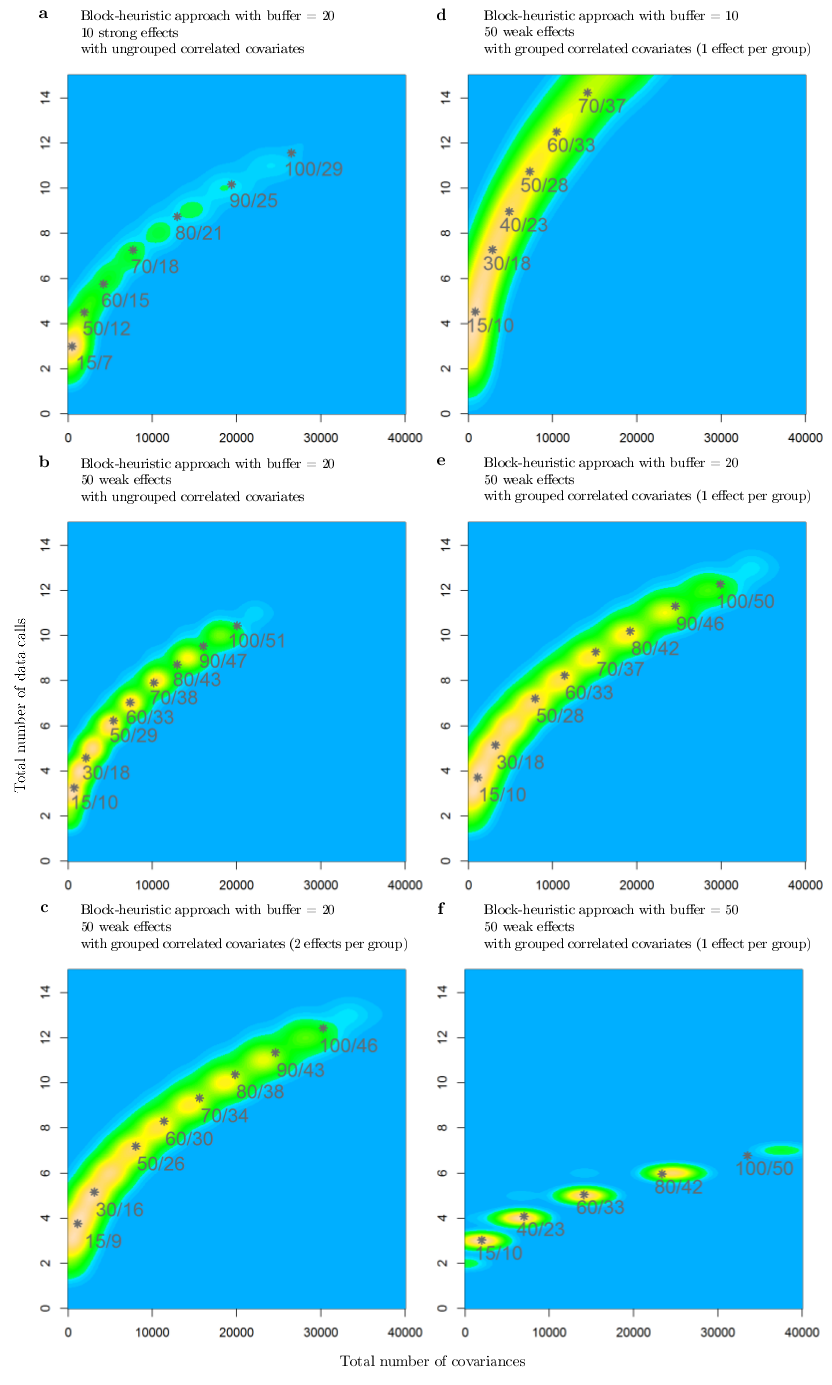}
\caption{Comparison of heuristic distributed boosting results in different simulation scenarios with varying buffer sizes. In panels a to c, we varied the simulation scenarios with a fixed buffer size of 20, and in panels d to f varied the buffer size in a fixed simulation setting. The distribution of the total number of called covariances vs. the total number of data calls is calculated marginally over $100$ boosting steps and $100$ simulations. A blue color corresponds to zero, and the more yellow the color, the more often the specific combination occurs. The grey stars represent the mean number of data calls and called covariances for specific boosting step numbers (before the slash) and corresponding model sizes (after the slash).}
\label{Figure_block}
\end{figure}

In Figures \ref{Figure_heuristic} and \ref{Figure_block}, smoothed frequencies of combinations of the total number of data calls and the total number of called covariances over $100$ boosting steps and $100$ simulation runs are shown for the classic heuristic distributed boosting approach and the block-heuristic version. A blue color corresponds to zero, and more yellowish coloring indicates a combination which occurs more frequently. The grey stars represent the mean values for specific boosting step numbers (before the slash) and corresponding model sizes (after the slash), i.e., the number of included covariates. Figure \ref{Figure_heuristic} focuses on the comparison between the heuristic and the block-heuristic approaches in a simulation setting with $10$ covariates with a substantial effect, a moderate overall covariance, and a buffer of $20$ for the block-heuristic approach. Figure \ref{Figure_block} visualizes the performance of the block-heuristic approach in different simulation scenarios with varying buffer sizes. We show all results for $n=500$ individuals distributed over five cohorts. The results for different sample sizes and a varying number of cohorts do not differ strongly. 

The results of the heuristic distributed boosting approach (Figure \ref{Figure_heuristic}, panel a) indicate that this approach can reduce the number of covariates compared to the classic non-heuristic approach. Even if one performs $100$ boosting steps to achieve a model size of about $50$ covariates, one would need less than $10000$ covariate values on average. If one additionally considers the desired model size of $10$ covariates, less than $300$ covariate values on average are called. Consequently, the amount of data needed can be reduced by about $99\%$ compared to the standard non-heuristic distributed boosting approach. On the other hand, the heuristic increases the number of data calls. In the given simulation scenario the number of data calls is approximately twice as large as the number of included covariates. In other simulation scenarios, the factor differs, and the number of data calls can be comparable to the standard approach, but overall the number of data calls is increased.

The number and the size of the data calls in the block-heuristic approach strongly depend on the buffer size $w$. The larger the chosen buffer, the more data is called in one data call. Although some values might be called but never needed for the algorithm, the idea is that values may be needed later on are already transferred to the analysis server before they are needed, and thus the number of data calls can be reduced. In the simulation settings with $50$ weak effect bearing covariates and a grouped covariance structure with one effect per group (Figure \ref{Figure_block}, panel d to f), one can see that for $100$ boosting steps the number of called covariances increases from about $30000$ (buffer $w=10$) to about $40000$ (buffer $w=50$). On the other hand, the number of data calls is reduced to about $20$ or $7$, respectively. For a model size of $10$ covariates, the increase in the number of called covariances is negligible (remaining smaller than $2,500$), but the number of data calls is reduced to about $3-4$. On average, a buffer size of $w=20$ also requires about $3$ data calls to achieve a model size of $10$ covariates in the other simulation scenarios (Figure \ref{Figure_block}, panels a to c). With this buffer size the number of called covariances is increased compared to the heuristic version (Figure \ref{Figure_heuristic}), but compared to the standard distributed boosting approach, the amount of called data was still reduced by at least $90\%$ in all of the simulation settings considered. In our simulation setting, as expected, the buffer size $w$ does not strongly influence the results of the variable selection.

\section*{Discussion}
The discovery and validation of biomarkers requires large patient cohorts and is aided by pooled and joint data approaches across institutions. The use of regularized regression techniques for selecting potentially important cancer biomarkers when jointly analyzing data from different data sources is hindered by data protection constraints, e.g., governance restrictions and fear of loss of intellectual property, which make pooling of individual-level data difficult or impossible. By re-formulating the statistics needed to perform componentwise likelihood-based boosting, we have proposed a method adequate for this situation. This proposed approach is solely based on univariable effect estimates and pairwise covariances, i.e. aggregated data. As the re-formulation involves no approximations, the results of standard componentwise likelihood-based boosting using individual-level data and the proposed alternative are the same if the data are standardized globally. 

In the given data situation, in which data are distributed over several cohorts, it is possible that the data need to be standardized at each data site. The loss of performance due to the distribution process was only minimal in the simulation settings we considered, in particular if the number of cohorts was smaller than $10$. In addition, we showed empirically that the proposed method outperformed a standard univariable approach even for $20$ cohorts. The loss in prediction performance was negligible overall. 

Besides the fact that the full covariance matrix of all covariates might be too big to exchange without difficulty, one could argue that using the full covariance matrix risks reconstructing individual-level data as the number of shared values might exceed the number of originally collected values. To ameliorate this problem, we proposed using a heuristic approach. Instead of using the full covariance matrix, the heuristic approach can identify the most promising covariates and only needs these to perform the next step of the algorithm. Thereby, the amount of data required can be reduced by up to $99\%$. Unfortunately, the algorithm then needs to call for data more often, which might be problematic, e.g., if the data release is handled manually by the data sites. To address this problem, we propose an alternative heuristic approach which moderately increases the amount of called data. Instead of calling single values of the covariance matrix, it calls full blocks with an additionally added buffer $w$. From experience, a buffer size of $w=20$ results in about $3$ data calls for a model consisting of $10$ covariates, and reduces the amount of called data by at least $90\%$. As a consequence, the proposed algorithm is fast and feasible for practical use, allowing even exploratory analysis to be performed. We also investigated alternative block-heuristic versions, where the buffer was set to zero or only added in one direction of the block. These versions performed worse, and are not discussed here.

However, the presented method has some limitations. First, some information is lost if the data is standardized locally, although the performance loss is negligible in realistic settings with consortia of up to $10$ members. A larger number of members is also possible, but might lead to a greater difference between a global and a local standardization approach. On the other hand, standardization at each data site indirectly allows for different intercepts and thus prevalences, which is to be expected when combining different cohorts. The algorithm can also easily be extended to allow for different effect estimates, meaning that the combination of different cohorts is only used for variable selection and the analysis can be adjusted for baseline differences. Moreover, the choice of the buffer size $w$ is arbitrary. Nevertheless, in general this choice only influences the number and size of the data calls whereas the variable selection itself is stable. Only in situations where the heuristic itself should not be used, i.e. when highly correlated variables are expected to have opposing directions of effects, might the results differ. Furthermore, the simulation study only reflected rather simple settings. In particular, we did not vary the effect sizes, the number of individuals per cohort or the covariate distributions. Additionally, we only considered a binary endpoint although our proposed approach formally uses linear regression with a continuous outcome, but as the method is exact for the continuous outcomes, the performance loss due to the distribution of the data would be smaller.  

Until now only a limited number of methods have been adapted to distributed data under data protection constraints. This paper provides evidence that even machine learning approaches, which currently heavily depend on individual-level data, can be adapted to these settings. In order to achieve this, existing methods such as LASSO or random forest should be re-checked for potential ways to re-formulate the underlying mathematical calculations in such a way that non-disclosive aggregated data are sufficient. Such methods can increase data protection and grant access to data sources which would not otherwise be publicly available. The proposed method can identify covariates associated with a binary endpoint in large scale omics data and can construct statistical models solely based on aggregated, distributed data. By additionally using a block-heuristic approach, the algorithm is easily applicable in consortia. A readily accessible implementation of the proposed algorithm, using the DataSHIELD framework, is available as a package from \texttt{https://github.com/danielazoeller/ds\_DistributedBoosting.jl} using the Julia language. An extension to (competing) event time analysis via direct binomial regression is possible, e.g., via pseudo values.

\subsection*{Conclusions}
Statistical methods based on aggregated data are promising techniques to enable joint analysis of data under data protection constraints without pooling. We proposed a regularized multivariable regression approach for building prediction signatures using automatic variable selection based on summary statistics, namely univariable effect estimates and pairwise covariances. In consortia with less than ten participating sites, we can identify effect bearing covariates to a comparable extent as when pooling individual-level data. In large consortia, we would expect only a small performance loss. Additionally, we proposed a heuristic variant that can effectively reduce the number and size of data calls to enable a manual data release at participating data sites. Thus, we consider (heuristic) distributed boosting to be a valuable approach to support joint analysis of several data sources under data protection constraints. Importantly, our multivariable regression approach will foster and accelerate the discovery of new cancer biomarkers by facilitating the cross-institutional analysis of patient cohorts without limitations imposed by patient data or intellectual property requirements, thereby improving future cancer therapy and care. Other existing statistical and machine learning approaches should be evaluated for similar potential to re-formulate the underlying mathematical basis in such a way that non-disclosive aggregated data are sufficient to ensure data protection.

\section*{Acknowledgements}
  The authors like to thank Marcel Schniedermann for computational support and in particular for compiling the figures on heuristic distributed boosting. This work contains parts of the Ph.D. thesis of Daniela Z\"oller.

\vspace*{1pc}

\section*{Conflict of Interest}

\noindent {\it{The authors have declared no conflict of interest.}}

\section*{Funding}
This research was funded by GEnder-Sensitive Analyses of mental health trajectories and implications for prevention: A multi-cohort consortium (GESA) of German Federal Ministry of Education and Research (BMBF) grant number FKZ 01GL1718A (D.Z.).

\section*{}
\bibliographystyle{plain}
\bibliography{References}

\begin{thebibliography}{10}

\bibitem{Andersen2010}
Per~Kragh Andersen and M.~{Pohar Perme}.
\newblock {Pseudo-observations in survival analysis}.
\newblock {\em Statistical Methods in Medical Research}, 19(1):71--99, 2010.

\bibitem{Ayers2010}
Kristin~L Ayers and Heather~J Cordell.
\newblock {SNP selection in genome-wide and candidate gene studies via
  penalized logistic regression}.
\newblock {\em Genetic epidemiology}, 34(8):879--891, 2010.

\bibitem{5496140}
Juan~Andrés Bazerque, Gonzalo Mateos, and Georgios~B. Giannakis.
\newblock Distributed lasso for in-network linear regression.
\newblock In {\em 2010 IEEE International Conference on Acoustics, Speech and
  Signal Processing}, pages 2978--2981, 2010.

\bibitem{Binder2013}
Harald Binder, Axel Benner, L~Bullinger, and M~Schumacher.
\newblock {Tailoring sparse multivariable regression techniques for prognostic
  single-nucleotide polymorphism signatures}.
\newblock {\em Statistics in Medicine}, 32(10):1778--1791, 2013.

\bibitem{Byrne2018}
Julianne Byrne, Desiree Grabow, Helen Campbell, Kylie O'Brien, Stefan Bielack,
  {Antoinette Am} Zehnhoff-Dinnesen, Gabriele Calaminus, Leontien Kremer,
  Thorsten Langer, {Marry M.} {van den Heuvel-Eibrink}, Eline {van Dulmen-den
  Broeder}, Katja Baust, Andrea Bautz, {J{\"o}rn D.} Beck, Claire Berger,
  Harald Binder, Anja Borgmann-Staudt, Linda Broer, Holger Cario, Leonie
  Casagranda, Eva Clemens, Dirk Deuster, Andrica {de Vries}, Uta Dirksen,
  {Jeanette Falck} Winther, Sophie Fossa, Anna Font-Gonzalez, Victoria
  Grandage, Riccardo Haupt, Stefanie Hecker-Nolting, Lars Hjorth, Melanie
  Kaiser, Line Kenborg, Tomas Kepak, Katerina Kepakova, {Lisbeth E.} Knudsen,
  Maryna Krawczuk-Rybak, Jarmila Kruseova, {Claudia E.} Kuehni, Marina
  Kunstreich, Rahel Kuonen, Herwig Lackner, Alison Leiper, {Erik A. H.}
  Loeffen, Ales Luks, Dalit Modan-Moses, Renee Mulder, Ross Parfitt, {Norbert
  W.} Paul, Andreas Ranft, Ellen Ruud, Ralph Schilling, Claudia Spix, Joanna
  Stefanowicz, Gabriele Strauss, {Andre G.} Uitterlinden, Marleen {van den
  Berg}, Anne-Lotte {van der Kooi}, Marloes {van Dijk}, Flora {van Leeuwen},
  Oliver Zolk, Daniela Zoeller, and Peter Kaatsch.
\newblock Pancarelife: The scientific basis for a european project to improve
  long-term care regarding fertility, ototoxicity and health-related quality of
  life after cancer occurring among children and adolescents.
\newblock {\em European Journal of Cancer}, 103:227--237, 2018.

\bibitem{Efron2004}
Bradley Efron, Trevor Hastie, Iain Johnstone, and Robert Tibshirani.
\newblock {Least angle regression}.
\newblock {\em The Annals of Statistics}, 32(2):407--499, 2004.

\bibitem{Emam2013}
Khaled~El Emam, Saeed Samet, Luk Arbuckle, Robyn Tamblyn, and Craig Earle.
\newblock {A secure distributed logistic regression protocol for the detection
  of rare adverse drug events}.
\newblock {\em J Am Med Inform Assoc}, 20:453--461, 2013.

\bibitem{Fecher2015}
Benedikt Fecher, Sascha Friesike, and Marcel Hebing.
\newblock {What drives academic data sharing?}
\newblock {\em PLOS ONE}, 10(2), 2015.

\bibitem{Gaye2014a}
Amadou Gaye, Yannick Marcon, Julia Isaeva, P.~LaFlamme, Andrew Turner, Elinor~M
  Jones, Joel Minion, Andrew~W Boyd, Christopher~J Newby, M.-L. Nuotio, Rebecca
  Wilson, Oliver Butters, B.~Murtagh, Ipek Demir, Dany Doiron, Lisette
  Giepmans, Susan~E Wallace, I.~Budin-Ljosne, C.~{Oliver Schmidt}, P.~Boffetta,
  M.~Boniol, Maria Bota, Kim~W Carter, N.~DeKlerk, Chris Dibben, R.~W. Francis,
  T.~Hiekkalinna, K.~Hveem, K.~Kvaloy, Sean Millar, Ivan~J Perry, Annette
  Peters, C.~M. Phillips, F.~Popham, G.~Raab, E.~Reischl, N.~Sheehan,
  M.~Waldenberger, M.~Perola, E.~van~den Heuvel, John Macleod, Bartha~M
  Knoppers, R.~P. Stolk, I.~Fortier, J.~R. Harris, B.~H. Woffenbuttel, M.~J.
  Murtagh, V.~Ferretti, and P.~R. Burton.
\newblock {DataSHIELD: taking the analysis to the data, not the data to the
  analysis}.
\newblock {\em International Journal of Epidemiology}, 43(6):1929--1944, 2014.

\bibitem{Graw2009}
Frederik Graw, Thomas~Alexander Gerds, and Martin Schumacher.
\newblock {On pseudo-values for regression analysis in competing risks models}.
\newblock {\em Lifetime Data Analysis}, 15(2):241--255, 2009.

\bibitem{Hastie2007}
Trevor Hastie, Jonathan Taylor, Robert Tibshirani, and Guenther Walther.
\newblock Forward stagewise regression and the monotone lasso.
\newblock {\em Electron. J. Statist.}, 1:1--29, 2007.

\bibitem{He2016a}
Qianchuan He, Hao~Helen Zhang, Christy~L Avery, and D.~Y. Lin.
\newblock {Sparse meta-analysis with high-dimensional data}.
\newblock {\em Biostatistics}, 17(2):205--220, 2016.

\bibitem{Hinderer2017}
Marc Hinderer, Martin Boeker, Sebastian~A Wagner, Harald Binder, Frank Ückert,
  Stephanie Newe, Jan~L Hülsemann, Michael Neumaier, Carmen Schade-Brittinger,
  Till Acker, Hans-Ulrich Prokosch, and Brita Sedlmayr.
\newblock The experience of physicians in pharmacogenomic clinical decision
  support within eight german university hospitals.
\newblock {\em Pharmacogenomics}, 18(8):773--785, 2017.

\bibitem{Jones2012}
E.~M. Jones, N.~A. Sheehan, N.~Masca, S.~E. Wallace, M.~J. Murtagh, and P.~R.
  Burton.
\newblock {DataSHIELD - shared individual-level analysis without sharing the
  data: a biostatistical perspective}.
\newblock {\em Norsk Epidemiologi}, 21(2):231--239, 2012.

\bibitem{leng2022federated}
Xinlin Leng, Chenxu Li, Weifeng Xu, Yuyan Sun, and Hongtao Wang.
\newblock Federated coordinate descent for privacy-preserving multiparty linear
  regression, 2022.

\bibitem{Li2016}
Wenfa Li, Hongzhe Liu, Peng Yang, and Wei Xie.
\newblock {Supporting regularized logistic regression privately and
  efficiently}.
\newblock {\em PLOS ONE}, 11(6):e0156479, 2016.

\bibitem{Liu2007}
Zhenqiu Liu, Feng Jiang, Guoliang Tian, Suna Wang, Fumiaki Sato, Stephen~J.
  Meltzer, and Ming Tan.
\newblock {Sparse logistic regression with Lp penalty for biomarker
  identification}.
\newblock {\em Statistical Applications in Genetics and Molecular Biology},
  6(1):6, 2007.

\bibitem{Lu2015}
Chia-lun Lu, Shuang Wang, Zhanglong Ji, Yuan Wu, Li~Xiong, Xiaoqian Jiang, and
  Lucila Ohno-machado.
\newblock {WebDISCO: a web service for distributed cox model learning without
  patient-level data sharing}.
\newblock {\em J Am Med Inform Assoc}, 22:1212--1219, 2015.

\bibitem{Pohlman2003}
John~T Pohlmann and Dennis~W Leitner.
\newblock A comparison of ordinary least squares and logistic regression.
\newblock {\em Ohio Journal of Science}, 103(5):118--125, 2003.

\bibitem{Prokosch2018}
Hans-Ulrich Prokosch, Till Acker, Johannes Bernarding, Harald Binder, Martin
  Boeker, Melanie Boerries, Philipp Daumke, Thomas Ganslandt, J{\"u}rgen
  Hesser, Gunther H{\"o}ning, Michael Neumaier, Kurt Marquardt, Harald Renz,
  Hermann-Josef Rothk{\"o}tter, Carmen Schade-Brittinger, Paul Schm{\"u}cker,
  J{\"u}rgen Sch{\"u}ttler, Martin Sedlmayr, Hubert Serve, Keywan Sohrabi, and
  Holger Storf.
\newblock Miracum: medical informatics in research and care in university
  medicine.
\newblock {\em Methods of information in medicine}, 57(S 01):e82--e91, 2018.

\bibitem{Schulz2019}
Jessica Schulz, Petros Takousis, Inken Wohlers, Ivie~O.G. Itua, Valerija
  Dobricic, Gerta Rücker, Harald Binder, Lefkos Middleton, John~P.A.
  Ioannidis, Robert Perneczky, Lars Bertram, and Christina~M. Lill.
\newblock Meta-analyses identify differentially expressed micrornas in
  parkinson's disease.
\newblock {\em Annals of Neurology}, 85(6):835--851, 2019.

\bibitem{Shafi2019}
Adib Shafi, Tin Nguyen, Azam Peyvandipour, Hung Nguyen, and Sorin Draghici.
\newblock A multi-cohort and multi-omics meta-analysis framework to identify
  network-nased gene signatures.
\newblock {\em Frontiers in genetics}, 10(159), 2019.

\bibitem{Simmons2016a}
Sean Simmons and Bonnie Berger.
\newblock {Realizing privacy preserving genome-wide association studies}.
\newblock {\em Bioinformatics}, 32(9):1293--1300, 2016.

\bibitem{Tutz2006a}
Gerhard Tutz and Harald Binder.
\newblock {Generalized additive modeling with implicit variable selection by
  likelihood-based boosting}.
\newblock {\em Biometrics}, 62(4):961--971, 2006.

\bibitem{Tutz2007}
Gerhard Tutz and Harald Binder.
\newblock {Boosting ridge regression}.
\newblock {\em Computational Statistics {\&} Data Analysis}, 51(12):6044--6059,
  2007.

\bibitem{Wegscheider2017}
Karl Wegscheider and Tim Friede.
\newblock {Do we consent to rules of consent and confidentiality?}
\newblock {\em Biometrical Journal}, 59(2):235--239, mar 2017.

\bibitem{Williams2017}
Garrath Williams and Iris Pigeot.
\newblock {Consent and confidentiality in the light of recent demands for data
  sharing}.
\newblock {\em Biometrical Journal}, 59(2):240--250, 2017.

\bibitem{Zenker2022}
Sven Zenker, Daniel Strech, Kristina Ihrig, Roland Jahns, Gabriele M\"{u}ller,
  Christoph Schickhardt, Georg Schmidt, Ronald Speer, Eva Winkler,
  Sebastian~Graf von Kielmansegg, and Johannes Drepper.
\newblock Data protection-compliant broad consent for secondary use of health
  care data and human biosamples for (bio)medical research: Towards a new
  german national standard.
\newblock {\em Journal of Biomedical Informatics}, 131:104096, July 2022.

\bibitem{Zhu2019}
Ligeng Zhu, Zhijian Liu, and Song Han.
\newblock {\em Deep Leakage from Gradients}.
\newblock Curran Associates Inc., Red Hook, NY, USA, 2019.

\bibitem{Zhu2017}
X.~{Zhu}, H.~{Suk}, H.~{Huang}, and D.~{Shen}.
\newblock Low-rank graph-regularized structured sparse regression for
  identifying genetic biomarkers.
\newblock {\em IEEE Transactions on Big Data}, 3(4):405--414, 2017.

\bibitem{Zoller2015a}
Daniela Z{\"{o}}ller, Irene Schmidtmann, Arndt Weinmann, Thomas~A. Gerds, and
  Harald Binder.
\newblock {Stagewise pseudo-value regression for time-varying effects on the
  cumulative incidence}.
\newblock {\em Statistics in Medicine}, 35(7):1144--1158, 2016.

\end{thebibliography}
\end{document}